\documentclass[sigconf,nonacm]{acmart}%anonymous

\usepackage{booktabs} % For formal tables

\usepackage{multirow}
\usepackage{afterpage}

\setcopyright{acmcopyright}

\copyrightyear{2021} 
\acmYear{2021} 
\setcopyright{acmcopyright}\acmConference[GECCO '21 Companion]{2021 Genetic and Evolutionary Computation Conference Companion}{July 10--14, 2021}{Lille, France}
\acmBooktitle{2021 Genetic and Evolutionary Computation Conference Companion (GECCO '21 Companion), July 10--14, 2021, Lille, France}
\acmPrice{15.00}
\acmDOI{10.1145/3449726.3463137}
\acmISBN{978-1-4503-8351-6/21/07}

\begin{document}

\title{Black-box adversarial attacks using Evolution Strategies}
% \titlenote{}
% \subtitle{}
% \subtitlenote{}

\author{Hao Qiu}
\affiliation{%
%Department of Information Engineering and Computer Science, 
  \institution{University of Trento}
  \city{Trento}%Povo
  \country{Italy}
  %\streetaddress{Via Sommarive 9}
  %\postcode{38123}
}
\orcid{}
\email{qiuhaosai@gmail.com}

\author{Leonardo Lucio Custode}
\affiliation{%
%Department of Information Engineering and Computer Science, 
  \institution{University of Trento}
  \city{Trento}%Povo
  \country{Italy}
  %\streetaddress{Via Sommarive 9}
  %\postcode{38123}
}
\orcid{0000-0002-1652-1690}
\email{leonardo.custode@unitn.it}

\author{Giovanni Iacca}
\affiliation{%
%Department of Information Engineering and Computer Science, 
  \institution{University of Trento}
  \city{Trento}%Povo
  \country{Italy}
  %\streetaddress{Via Sommarive 9}
  %\postcode{38123}
}
\orcid{0000-0001-9723-1830}
\email{giovanni.iacca@unitn.it}

%\shortauthors{%Qiu et al.}

\begin{abstract}
In the last decade, deep neural networks have proven to be very powerful in computer vision tasks, starting a revolution in the computer vision and machine learning fields. However, deep neural networks, usually, are not robust to perturbations of the input data. In fact, several studies showed that slightly changing the content of the images can cause a dramatic decrease in the accuracy of the attacked neural network. Several methods able to generate adversarial samples make use of gradients, which usually are not available to an attacker in real-world scenarios. As opposed to this class of attacks, another class of adversarial attacks, called black-box adversarial attacks, emerged, which does not make use of information on the gradients, being more suitable for real-world attack scenarios. In this work, we compare three well-known evolution strategies on the generation of black-box adversarial attacks for image classification tasks. While our results show that the attacked neural networks can be, in most cases, easily fooled by all the algorithms under comparison, they also show that some black-box optimization algorithms may be better in ``harder'' setups, both in terms of attack success rate and efficiency (i.e., number of queries).
\end{abstract}


\begin{CCSXML}
<ccs2012>
   <concept>
       <concept_id>10010147.10010257</concept_id>
       <concept_desc>Computing methodologies~Machine learning</concept_desc>
       <concept_significance>500</concept_significance>
       </concept>
   <concept>
       <concept_id>10010147.10010178.10010224</concept_id>
       <concept_desc>Computing methodologies~Computer vision</concept_desc>
       <concept_significance>500</concept_significance>
       </concept>
   <concept>
       <concept_id>10010147.10010178.10010205</concept_id>
       <concept_desc>Computing methodologies~Search methodologies</concept_desc>
       <concept_significance>500</concept_significance>
       </concept>
 </ccs2012>
\end{CCSXML}

\ccsdesc[500]{Computing methodologies~Machine learning}
\ccsdesc[500]{Computing methodologies~Computer vision}
\ccsdesc[500]{Computing methodologies~Search methodologies}

\keywords{Adversarial attacks, evolution strategies, CMA-ES, neural networks}

\maketitle

%----------------------------------------------

\section{Introduction}
\label{sec:introduction}
The field of computer vision, in the last decade, had an impressive progress that enabled applications such as autonomous driving, medical applications and identification.
All of these progresses are due to the capabilities of deep artificial neural networks in processing raw data such as images.
While deep neural networks are able to recognize, with good accuracy, objects in an image, they usually suffer under adversarial attacks.
An adversarial attack is an image $\delta$ crafted in such a way that, given a correctly-classified image $x$, $x + \delta$ is misclassified.

There are two main classes of adversarial attacks: \textit{white-box} adversarial attacks and \textit{black-box} adversarial attacks.
White-box adversarial attacks can be seen as a simplified setting.
In fact, in this case, the adversary has full access to the neural network, and thus she can compute gradients on the classified samples.
This allows to find adversarial attacks by simply moving the image towards the direction that maximizes the gradient.
On the other hand, black-box adversarial attacks are more similar to real settings.
In fact, in this case the attacker has no access to the gradients.
Instead, she can only access the prediction or the output probabilities given by the model.
In this cases, gradient-based methods cannot be used.
Evolutionary computation, since it does not rely on the computation of the gradients, is therefore an appropriate tool for this task.

While previous works on black-box adversarial attacks by means of evolutionary algorithms focused on solving the problem of generating adversarial attacks in black-box settings \cite{Chen_2019_poba, alzantot2019genattack}, no work performed a comparison of various black-box optimization methods for these settings.
% Rew 2.1, 3.1
In fact, since in general different evolutionary algorithms can perform very differently on the same problem, choosing which algorithm should be used for black-box adversarial attacks is difficult. Thus, comparing various algorithms on this task can be of great interest in practical applications.

% Rew 2.1, 3.1
In order to assess how different evolutionary algorithms perform on the generation of adversarial examples for deep neural networks in black-box settings, in this work we compare three different evolution strategies (ES), namely: (1+1)-ES \cite{rechenberg1994evolutionsstrategie}, Natural Evolution Strategies \cite{wierstra2008natural} and the original version of CMA-ES \cite{hansen_1996_adapting, hansen06thecma}. We decided to focus our investigation on these three variants of ES for three main reasons. First of all, as shown in \cite{salimans2017evolution} and further discussed in \cite{such2017deep}, ES can rival backpropagation-based algorithms in deep reinforcement learning (RL) problems and as such it has recently attracted research attention also in the deep learning community. Secondly, ES can be essentially considered a gradient-based algorithm, since it performs a stochastic gradient descent based on a finite-difference approximation of the gradient \cite{such2017deep}. As such, it is worth investigating how ES can deal with a task such as the generation of adversarial examples that is typically tackled by (explicit) gradient-based methods. Lastly, we chose for our analysis two ES variants configured as population-less (i.e., handling one solution at a time), and CMA-ES, which is configured as a population-based algorithm and is considered nowadays the state-of-the-art in evolutionary optimization. Thus, we aim at finding if using a population rather than a single solution can provide a benefit on the task at hand.

The rest of the paper is structured as follows.
In the next section, we make a short overview of the recent developments in the field of adversarial attacks.
In Section \ref{sec:method}, we describe the details of the methods which are common to the three evolutionary algorithms. Then, in Section \ref{sec:results} we present the experimental results and discuss the performance of the three algorithms. Finally, in Section \ref{sec:conclusions} we present the conclusions of this work.

%----------------------------------------------

\section{Related work}
\label{sec:related_work}

\begin{figure*}[t!]
    \centering
    \includegraphics[width=0.94\textwidth]{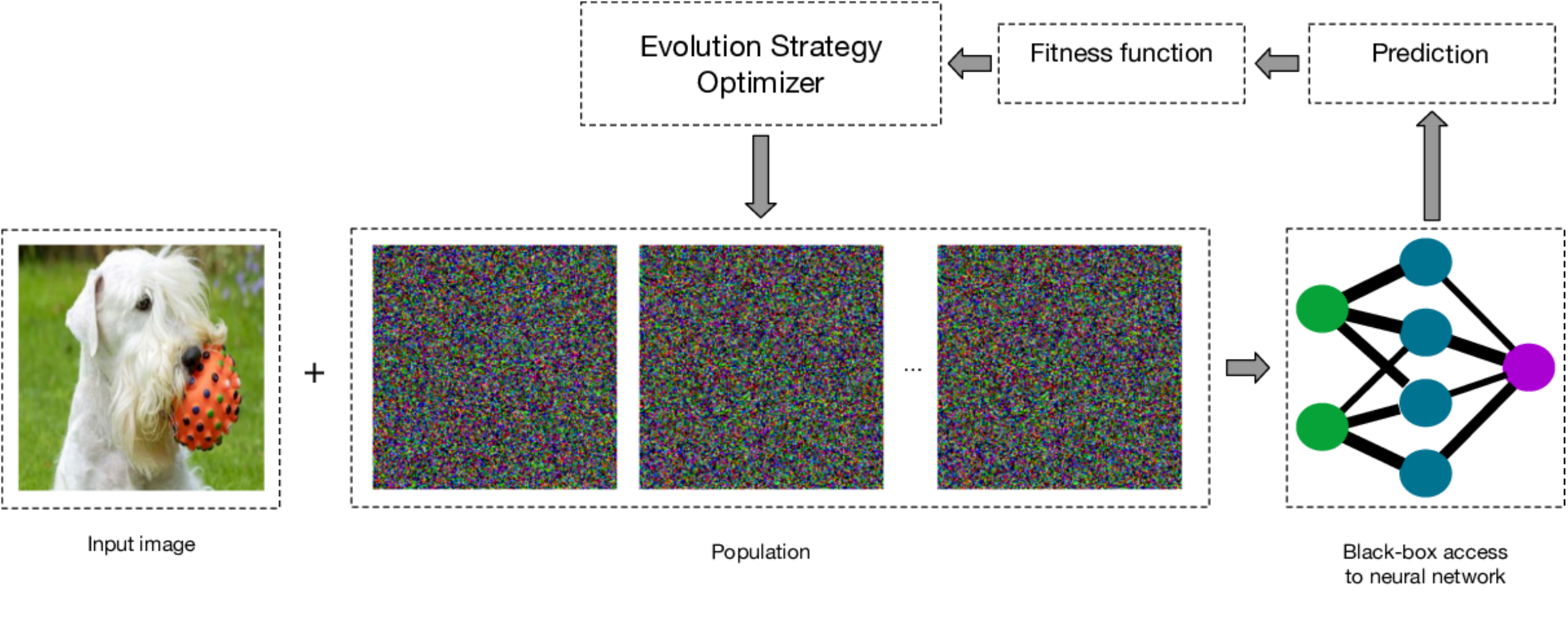}
    \caption{Optimization process used in our experiments to generate adversarial attacks.}
    \label{fig:process}
\end{figure*}

In the last years, the field of adversarial attacks has seen a number of important developments. 

Szegedy et al. \cite{szegedy2013intriguing} discovered that, while neural networks were able to reach good performance in image classification tasks, small perturbations of the input could cause them to misclassify samples that would be otherwise be classified correctly.
Successively, Goodfellow et al. \cite{goodfellow2014explaining} proposed an efficient gradient-based method to generate adversarial attacks.
In \cite{papernot2016limitations}, the authors addressed the generation of adversarial attacks in limited $L^0$-``norm'' scenarios, i.e., crafting adversarial samples by modifying only few pixels of the image.
Carlini and Wagner \cite{carlini2017towards} proposed a method to generate adversarial attacks that, by constraining the perturbation according to several metrics, were able to fool even neural networks that were trained on adversarial examples.
In \cite{moosavi2016deepfool}, the authors presented a method that generates adversarial examples by iteratively pushing the images outside the correct decision region.
Baluja and Fischer \cite{baluja2017adversarial} trained a neural network to generate adversarial examples for other neural networks.

All the approaches described before are designed for white-box settings, i.e., they require the attacker to have access to the model.
Other approaches, on the other hand, focused on black-box settings, i.e., settings in which the attacker knows only the logits of the prediction.

Su et al. \cite{su2019one} proposed an approach based on the differential evolution algorithm \cite{storn1997differential} to evolve adversarial examples that involve a single pixel in an image. 
In \cite{chen_2017_zoo}, the authors defined a special-purpose loss function and used it to estimate the gradients.
Ilyas et al. \cite{ilyas2018black} proposed a variant of the natural evolution strategies \cite{wierstra2008natural} to generate adversarial attacks.
In \cite{cheng2020signopt}, the authors introduced a method to generate black-box adversarial attacks even in stricter cases, i.e., when the attacker has access only to the predicted class.
Bhagoji et al. \cite{ferrari_practical_2018} proposed a loss function based on logits to estimate the gradients. Furthermore, they presented a query reduction technique to reduce the number of queries to obtain a successful attack.
Narodytska et al. \cite{narodytska2016perturbations} proposed a greedy algorithm for the generation of black-box adversarial attacks.
In \cite{narodytska_2017_attacks}, the authors made use of local search to generate black-box adversarial attacks.
Brendel et al. \cite{brendel2018decisionbased} proposed an algorithm to produce adversarial attacks based on an initial random image that is moved near the image that has to be attacked, leading to an adversarial attack that lies near the original image while being in a different decision region.
In \cite{Chen_2019_poba, alzantot2019genattack} the authors proposed a genetic algorithm to evolve adversarial perturbations.
Li et al. \cite{li2019nattack} proposed a method to learn the probability density distribution of adversarial attacks. 
In \cite{guo2019simple} the authors generated adversarial samples by generating noise in a direction obtained by analyzing orthonormal vectors of the image vector space.

Finally, some works \cite{kurakin_2017_adversarial, xu_adversarial_2019} have shown that it is possible to perform adversarial attacks even in real-world scenarios with physical objects.
For a more thorough review of the state of the art, the reader can refer to \cite{zhang_survey_2021, bhambri_survey_2020, qiu_review_2019, akhtar_threat_2018}.
% In this section, we describe the problem faced and make a short overview of the recent developments of the field, with the focus on black-box attack scenarios.

% % The problem of generating an adversarial attacks can be described as follows.
% % Given a neural network $f$ and a sample $x$, the problem consists in finding a perturbation $\delta$ (usually with some constraint, e.g. the maximum magnitude of $L^2$-norm, the $L^\infty$-norm or the $L^0$-``norm'') such that $f(x) = y_t$ and $f(x + \delta) = y_i$ with $y_i \ne y_t$.
% % If $y_i \ne y_t$ can be freely chosen among the other classes, we refer to an \textit{untargeted} attack.
% % Otherwise, when we constrain $y_i = y_{target}$, i.e., we want the neural network to classify $x$ as an instance of a particular class, we refer to a \textit{targeted} attack.

% Of course, in general, the adversary wants the adversarial attack to be undetectable to a human, so the problem of finding an adversarial attack can be though as:
% \[
%     \delta = \underset{\mid\delta'\mid < \varepsilon}{argmax}(L(x+\delta', y))
% \]
% where $L$ is a loss function associated to the output a neural network.

%----------------------------------------------

\section{Methods}

\label{sec:method}
In this work, we compare three different evolution strategies to generate adversarial samples in black-box settings.
The three algorithms we use are: (1+1)-ES \cite{rechenberg1994evolutionsstrategie}, Natural Evolution Strategies \cite{wierstra2008natural} and CMA-ES \cite{hansen_1996_adapting, hansen06thecma}. We omit, for brevity, the description of the three evolution strategies variants: for that, we refer the reader to the original papers.

The scheme of the process applied in our experiments to produce adversarial samples by means of these three evolutionary algorithms is shown in Figure \ref{fig:process}.

In the following, we describe the method used to generate the samples with the three evolutionary algorithms.

\subsection{Individual encoding}
Each individual is encoded as a $H \times W \times 3$ tensor, where $H$ and $W$ are the height and the width of the adversarial sample that is then upsampled to match the size of the image to attack (see Section \ref{sec:dim_red}), and 3 is the number of channels.

We constrain the $L^\infty$-norm of the generated adversarial perturbations.
To do so, we clip the values of the perturbations in $[-\varepsilon, \varepsilon]$.

\subsection{Fitness evaluation}
In the fitness evaluation phase, we compute a perturbation of the current image $x' = x + \delta$ and we use it to compute the goodness of the perturbation.
We refer to each time we assess the goodness of a perturbation of an adversarial attack as a \textit{query}.

\subsubsection{Untargeted attacks}
The fitness function used to assess the quality of an adversarial attack corresponds to the cross-entropy loss related to that sample with respect to the true label (i):
\[
F_u(x) = \mathcal{L}(x, y_i | y_i = 1) = - log(f_\theta^{(i)}(x))
\]
where $f_\theta^{(i)}(x)$ is the $i$-th output of the neural network parametrized by the parameters $\theta$, given the input $x$.
% Rew  2.3
In this case, instead of minimizing the cross-entropy loss (as done during training), we want to maximize it.

\subsubsection{Targeted attacks}
In this case, the fitness function used is:
\[
F_t(x) = -\mathcal{L}(x, y_t | y_t = 1) = log(f_\theta^{(t)}(x))
\]
where $t$ refers to the \textit{target} class.
Note that, in case of targeted attacks, we do not care about the correct class of the sample, but we only want to minimize the loss w.r.t. the target class.

When assessing the quality of an image, if the attack has success, the evolutionary process is terminated.

\subsection{Dimensionality reduction}\label{sec:dim_red}
In order to reduce the computational burden of the optimization process, we evolve adversarial samples that are smaller than the image to attack, and then we increase their size by means of nearest neighbor interpolation.
This means that, given an image $I$ and a scaling factor $s \geq 1$, we build another image $I'$ such that:
\[
I'(j\cdot s + k, i \cdot s + l) = I(j, i);\ \forall i, j;\ k, l \in [0, s[
\]

\begin{figure}[ht!]
    \centering
    \includegraphics[width=0.45\textwidth]{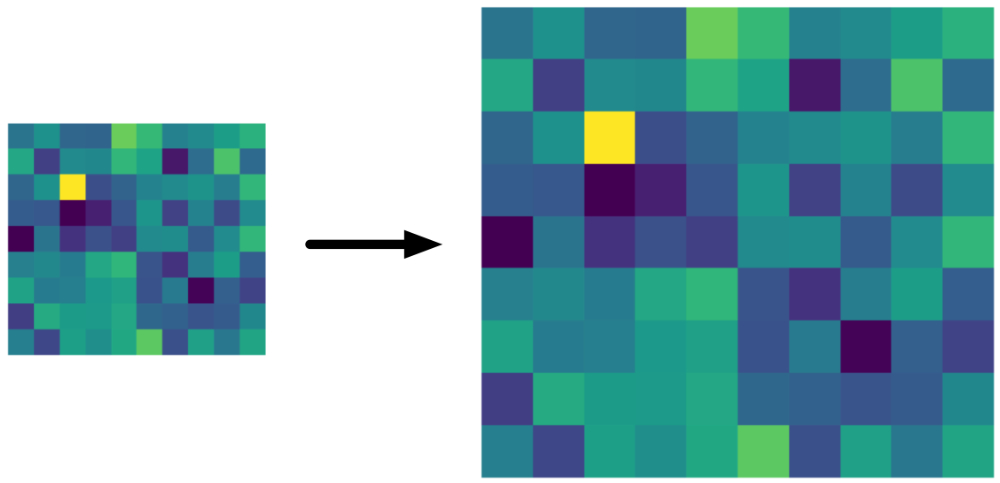}
    \caption{Example of upsampling using the nearest neighbor interpolation.}
    \label{fig:nn_interpolation}
\end{figure}

An example of upsampling using the nearest neighbor interpolation is shown in Figure \ref{fig:nn_interpolation}.

%----------------------------------------------

\section{Experimental results}
\label{sec:results}
In this section, we present the experimental setting and the results obtained.

\subsection{Experimental setting}
\subsubsection{Dataset}
We test the three evolutionary algorithms on neural networks that have been trained on the ImageNet dataset, that is a huge database composed of more than 14 million images distributed over more than $2 \cdot 10^4$ classes.
Each image is made of $224 \times 224$ pixels, encoded as a tensor in the RGB color space.

\subsubsection{Target neural networks}
Besides testing the three evolutionary algorithms on the ImageNet dataset, we also test them on three different neural networks, to better understand their performance in different scenarios.
The target neural networks are: VGG-16 \cite{simonyan2014very}, Inception-v3 \cite{szegedy2016rethinking} and ResNet-50 \cite{he2016deep}.

\subsubsection{Individual encoding}
As discussed earlier, the individuals are encoded as tensors of size $32\times32\times3$, that are then upsampled to $224\times224\times3$ tensors and added to the original image.

\subsubsection{Algorithm parametrization}
The parameters for the (1+1)-ES, NES and CMA-ES are shown in Table \ref{tab:params_es}.

\begin{table}[ht!]
    \centering
    \begin{tabular}{c|c|c} \hline
        \textbf{Algorithm} & \textbf{Parameter} & \textbf{Value} \\ \hline
        \multirow{3}{*}{(1+1)-ES} & Initialization strategy & $\mathcal{N}(0, 1)$ \\
        & Adaptation rule & 1/5 rule \\ 
        & Generations & 10000 \\ \hline
        \multirow{5}{*}{NES} & Initialization strategy & $\mathcal{N}(0, 1)$ \\
        & Population size & 1 \\
        & Step size & 1 \\
        & Learning rate & 0.05 \\ 
        & Generations & 10000 \\ \hline
        \multirow{3}{*}{CMA-ES} & Initialization strategy & $\mathcal{N}(0, 1)$ \\
        & Population size & 25 \\ 
        & Generations & 400 \\ \hline
    \end{tabular}
    \caption{Parameters used for the evolutionary algorithms employed in our experiments.}
    \label{tab:params_es}
\end{table}

\subsection{Experimental results}
Table \ref{tab:imagenet} shows the results obtained by using the three evolutionary algorithms in the untargeted setting on the three neural networks.
The maximum number of queries (for all the experiments) is set to $10^4$.
We set the maximum perturbation strength to $\varepsilon = 0.05$.

\begin{figure}[ht!]
    \centering
    \resizebox{0.47\textwidth}{!}{
        \includegraphics{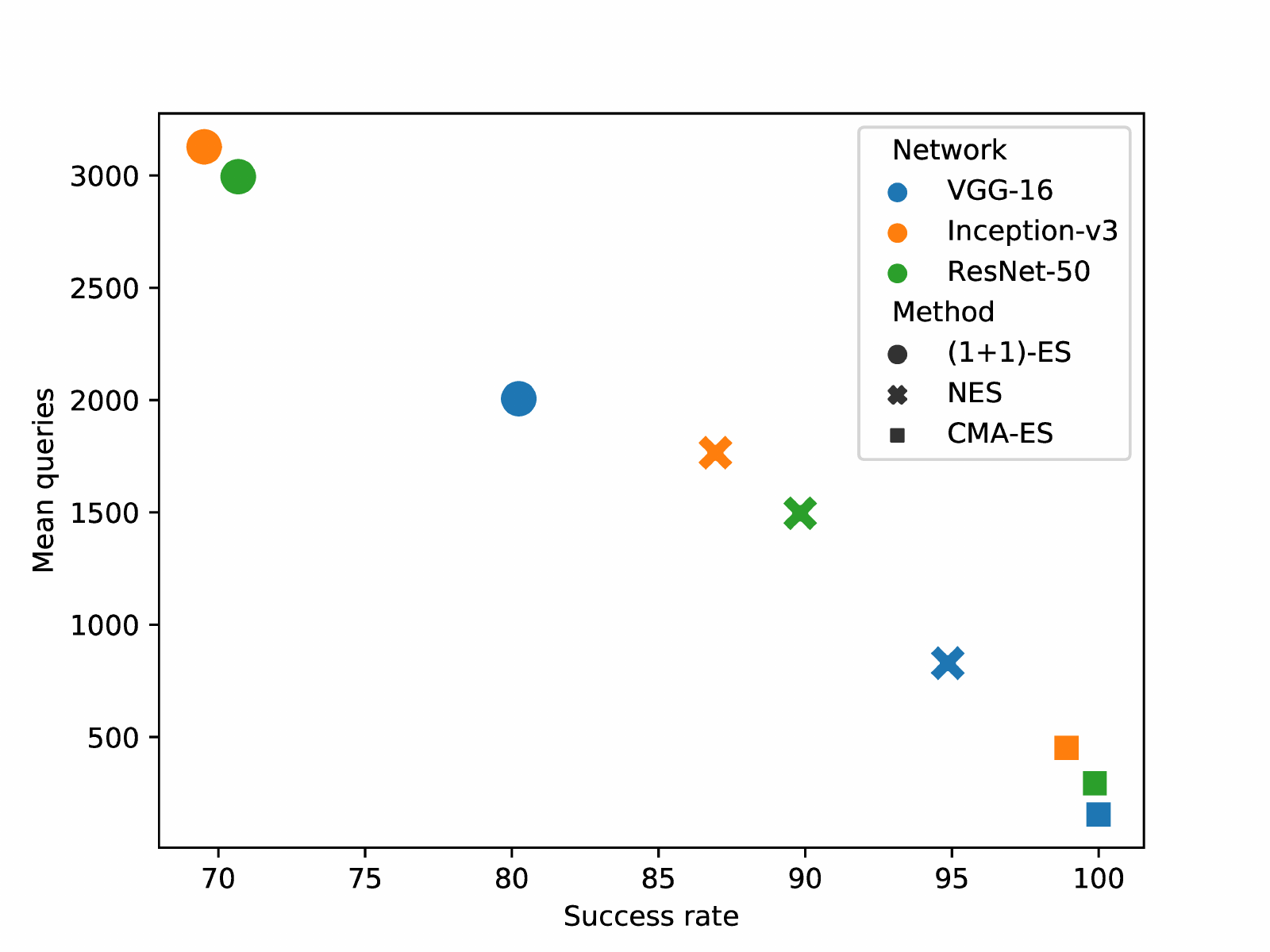}
    }
    \caption{Performance of the three evolutionary algorithms in terms of success rate and mean number of queries required while attacking the three neural networks ($\varepsilon = 0.05$).}
    \label{fig:success_queries_plot}
\end{figure}

Figure \ref{fig:success_queries_plot} graphically shows the attack success rate and the mean number of queries of the three algorithms on the three tested networks. It can be seen that CMA-ES ``dominates'' (as in Pareto domination) the other two algorithms in all the tested cases.

\begin{table*}[ht!]
    \centering
    \begin{tabular}{c|c|c|c|c} \hline
        \textbf{Neural Network} & \textbf{Algorithm} & \textbf{Success rate (\%)} & \textbf{Mean queries} & \textbf{Median queries} \\ \hline
        \multirow{3}{*}{VGG-16} & (1+1)-ES & 80.24 & 2006.00 (37.80) & 3.00 \\
        & NES & 94.86 & 828.90 (331.80) & 3.00 \\
        & CMA-ES & 100.00 & 155.80 (155.80) & 3.00 \\ \hline
        \multirow{3}{*}{Inception-v3} & (1+1)-ES & 69.52 & 3126.90 (114.00) & 32.00 \\
        & NES & 86.94 & 1765.50 (528.40) & 71.00 \\
        & CMA-ES & 98.91 & 451.20 (346.10) & 53.00 \\ \hline
        \multirow{3}{*}{ResNet-50} & (1+1)-ES & 70.68 & 2994.30 (88.50) & 17.00 \\
        & NES & 89.83 & 1496.10 (532.90) & 19.00 \\
        & CMA-ES & 99.87 & 294.20 (281.2) & 21.00 \\ \hline
    \end{tabular}
    \caption{Comparison of the three evolutionary algorithms with a maximum perturbation strength $\varepsilon = 0.05$ on the three neural networks pretrained on ImageNet. The values in parentheses represent the mean number of queries computed only on the successful attacks.}
    \label{tab:imagenet}
\end{table*}

From the data shown in Table \ref{tab:imagenet} and Figure \ref{fig:success_queries_plot}, we can observe that Inception-v3 seems to be more robust to the type of adversarial images produced in this work than VGG-16 and ResNet-50.
Moreover, we can observe that CMA-ES largely outperforms the other two algorithms on all the three networks, both in terms of success rate and mean number of queries. This might be due to the use of an actual population in CMA-ES (as opposed to the single-solution approach of the other two methods), which might favor a better exploration of the fitness landscape. However, further investigations will be needed to confirm this hypothesis. %Rew 2.2

\begin{figure}[ht!]
    \centering
    \includegraphics[width=0.47\textwidth]{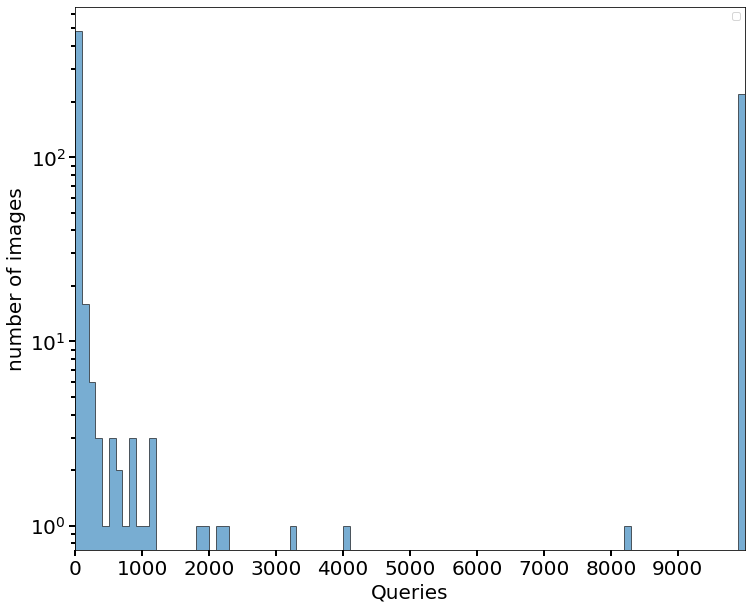}
    \caption{Distribution of the number of queries required to successfully attack the ResNet50 model by using (1+1)-ES.}
    \label{fig:queries_11}
\end{figure}

\begin{figure}[ht!]
    \centering
    \includegraphics[width=0.47\textwidth]{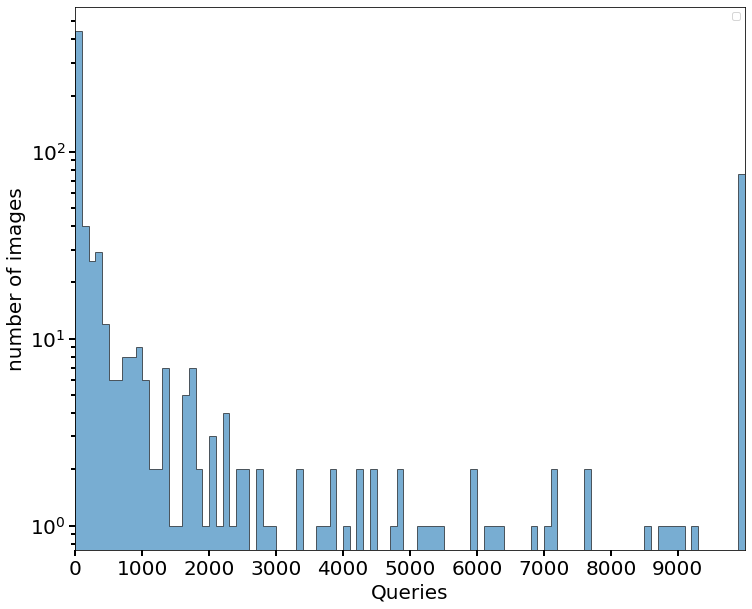}
    \caption{Distribution of the number of queries required to successfully attack the ResNet50 model by using NES.}
    \label{fig:queries_nes}
\end{figure}

\begin{figure}[ht!]
    \centering
    \includegraphics[width=0.47\textwidth]{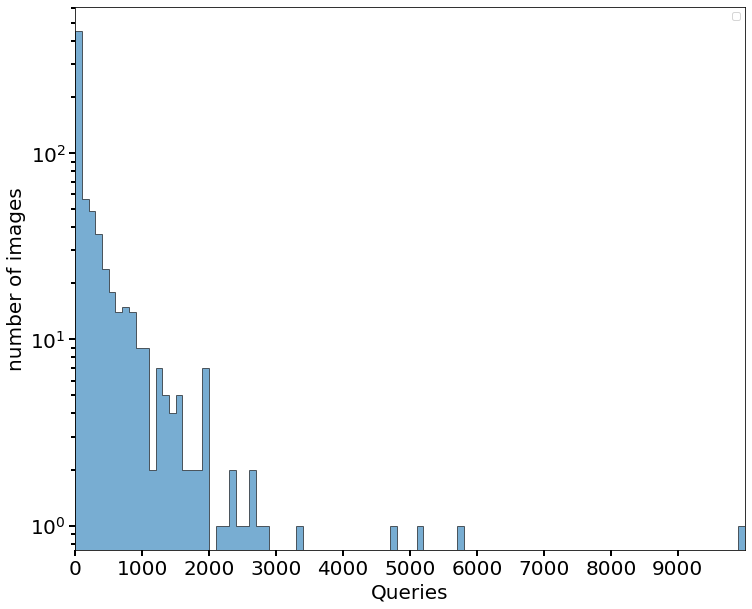}
    \caption{Distribution of the number of queries required to successfully attack the ResNet50 model by using CMA-ES.}
    \label{fig:queries_cmaes}
\end{figure}

On the other hand, it is interesting to note that a simple approach like (1+1)-ES consistently requires a smaller number of median queries.
This, and the fact that the mean number of queries in case of success is smaller than the one of CMA-ES, suggest that (1+1)-ES may be a good approach in certain cases (e.g. when the needed success rate is not too high, or when one can use only a limited amount of computing resources).  % Rew 2.6
Figures \ref{fig:queries_11}, \ref{fig:queries_nes}, \ref{fig:queries_cmaes} show the distribution of the queries performed while attacking the ResNet50 network for the (1+1)-ES, NES and CMA-ES, respectively.

To better understand how the maximum magnitude of the perturbation affects the adversarial attacks, we performed an additional test on the ResNet-50 using different perturbation strengths.
We performed this test on ResNet-50 because (from the results shown above) in terms of robustness to the attacks this network can be seen as an average case between VGG-16 and Inception-v3.
To do so, we tested the three evolutionary algorithms under varying perturbation strengths ($L^\infty$-norm) from $0.01$ to $0.09$, with steps of $0.02$.
The results are shown in Table \ref{tab:resnet-50-strength}.

As we can see from the table, even with small perturbations CMA-ES is able to obtain a very good success rate.
On the other hand, if the maximum perturbation is small, (1+1)-ES and NES are not able to achieve a satisfactory success rate.

Furthermore, higher perturbations ($\varepsilon \in [0.05, 0.09]$) show that even random noise is sufficient to obtain adversarial samples.
In fact, the median number of queries is so low that it means that even individuals in the initial generations (i.e., randomly generated) are able to fool the target neural network.
Nevertheless, also in this case CMA-ES proves to be superior to the other two evolutionary algorithms.
In fact, while all the algorithms obtain satisfactory success rates in (almost) all the cases, CMA-ES is the only one able to achieve a success rate of 100\%.


\begin{table*}[ht!]
    \centering
    \begin{tabular}{c|c|c|c|c} \hline
        \textbf{$L^\infty$-norm} & \textbf{Algorithm} & \textbf{Success rate (\%)}  & \textbf{Mean queries} & \textbf{Median queries} \\ \hline
        \multirow{3}{*}{0.01} & (1+1)-ES & 22.13 & 8009.60 (1007.80) & $10^4$ \\
        & NES & 35.70 & 6935.50 (1417.20) & $10^4$ \\
        & CMA-ES & 83.53 & 3218.80 (1881.80) & 1276.00 \\ \hline
        \multirow{3}{*}{0.03} & (1+1)-ES & 48.61 & 5245.10  (219.60) & $10^4$ \\
        & NES & 71.01 & 3522.50 (878.20) & 687.00 \\
        & CMA-ES & 99.60 & 701.00 (664.10) & 335.00 \\ \hline
        \multirow{3}{*}{0.05} & (1+1)-ES & 70.68 & 2994.30  (88.50) & 17.00 \\
        & NES & 89.83 & 1496.10 (532.90) & 19.00 \\
        & CMA-ES & 99.87 & 294.20 (281.20) & 21.00 \\ \hline
        \multirow{3}{*}{0.07} & (1+1)-ES & 86.42 & 1416.60 (68.90) & 4.00 \\
        & NES & 96.57 & 614.40 (281.40) & 3.00 \\
        & CMA-ES & 100.00 & 125.40 (125.40) & 3.00 \\ \hline
        \multirow{3}{*}{0.09} & (1+1)-ES & 95.25 & 494.90 (21.60) & 2.00 \\
        & NES & 98.94 & 250.40 (146.50) & 1.00 \\
        & CMA-ES & 100.00 & 59.70 (59.70) & 1.00 \\ \hline
    \end{tabular}
    \caption{Comparison of the three evolutionary algorithms on the generation of adversarial attacks for ResNet-50 under different maximum perturbation strengths. The values in parentheses represent the mean number of queries computed only on the successful attacks.}
    \label{tab:resnet-50-strength}
\end{table*}

\begin{figure}[ht!]
    \centering
    \includegraphics[width=0.47\textwidth]{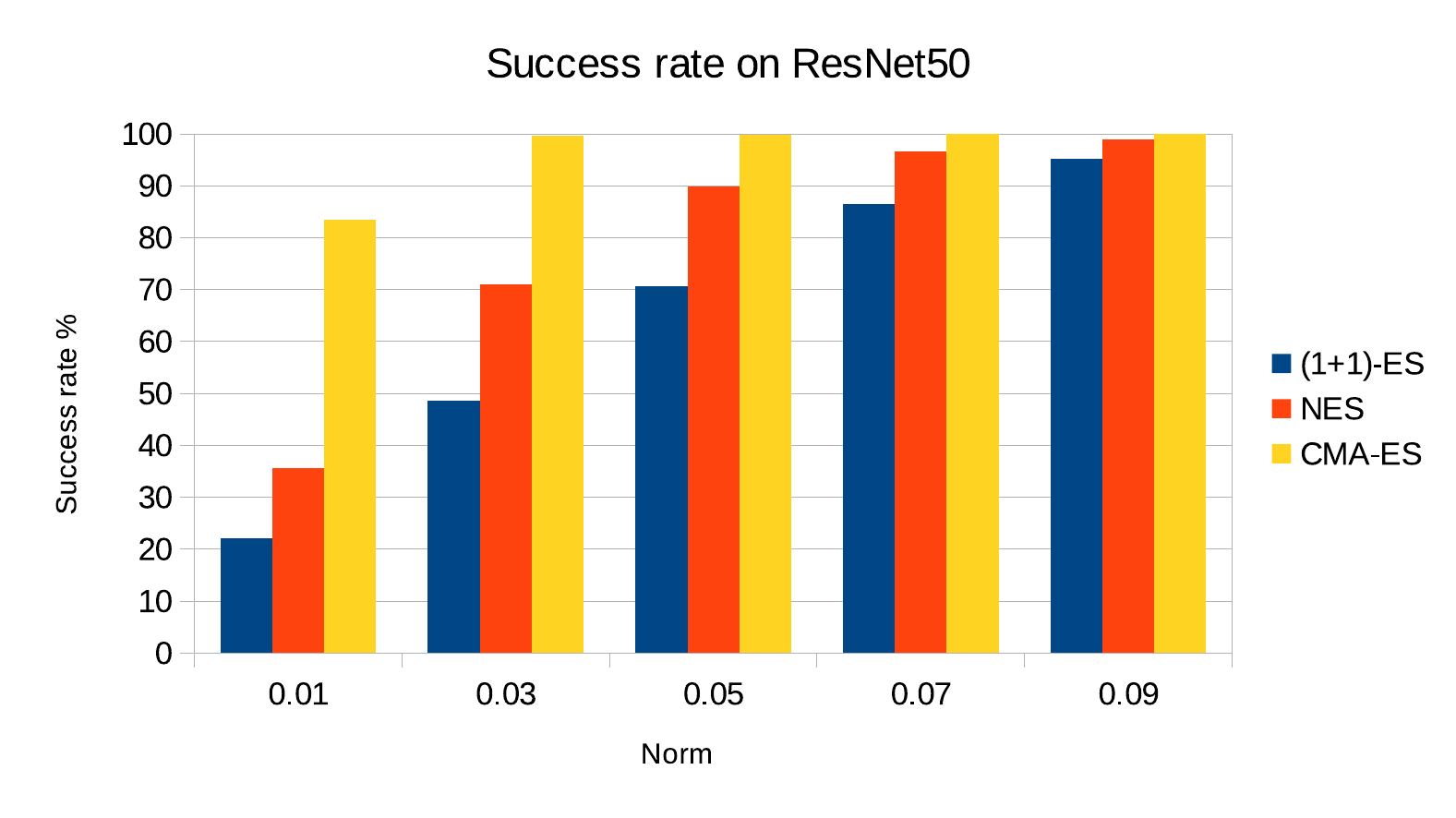}
    \caption{Success rate of the targeted attacks while using the three evolutionary algorithms to attack ResNet50 under different $L^\infty$-norms.}
    \label{fig:barsuccessrate}
\end{figure}

\begin{figure}[ht!]
    \centering
    \includegraphics[width=0.47\textwidth]{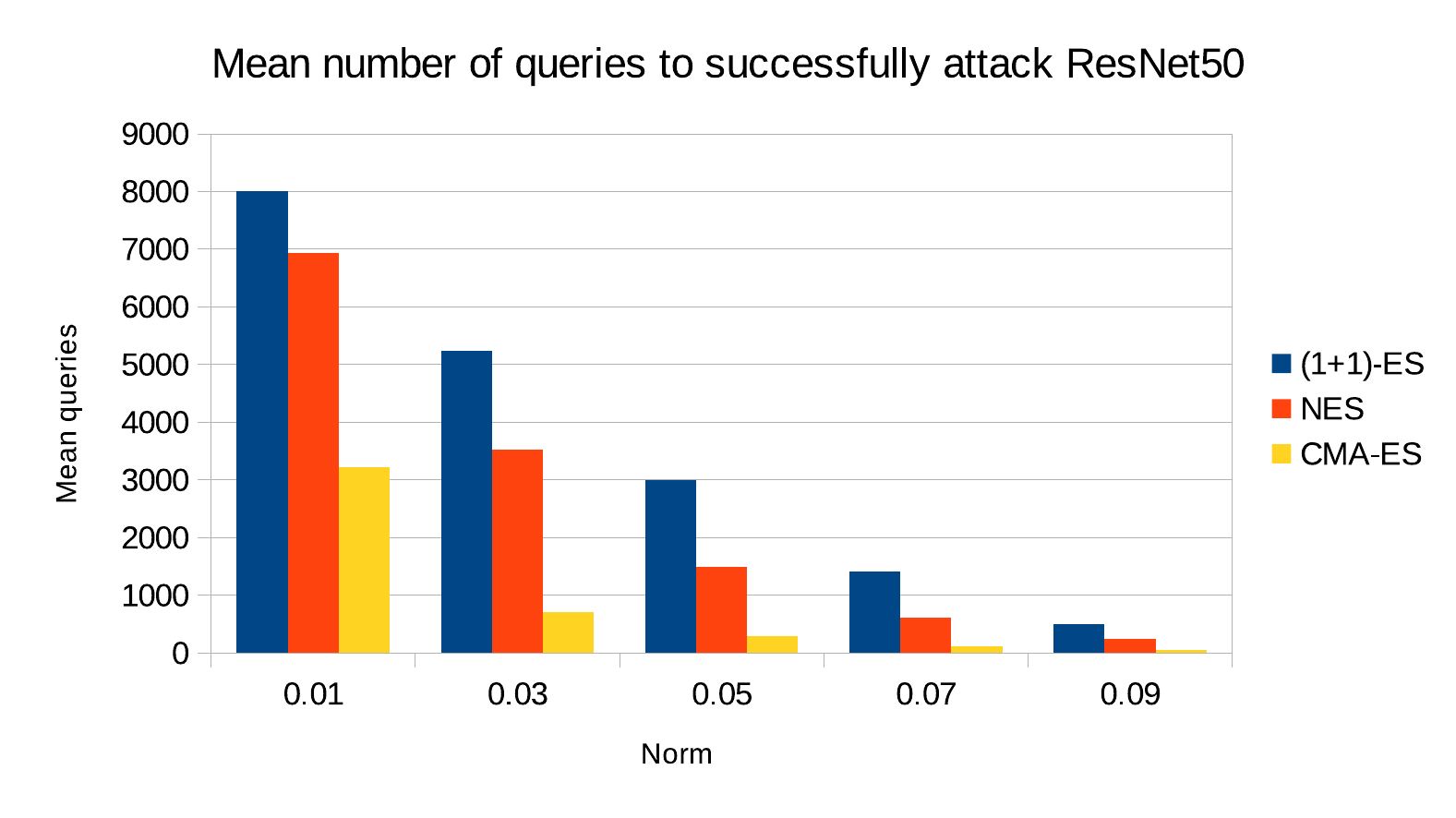}
    \caption{Mean number of queries needed to generate a successful targeted attack while using the three evolutionary algorithms to attack ResNet50 under different $L^\infty$-norms.}
    \label{fig:barmeanqueries}
\end{figure}

Finally, we tested the three algorithms on the generation of \textit{targeted} attacks for the ResNet-50 network.
The results of these experiments are presented in Table \ref{tab:targeted-resnet-50}.
The success rates under different $L^\infty$-norms are shown graphically in Figure \ref{fig:barsuccessrate}, while the mean number of queries is shown graphically in Figure \ref{fig:barmeanqueries}.

\begin{table}[ht!]
    \centering
        \begin{tabular}{c|c|c|c} \hline
            \textbf{Algorithm} & \textbf{SR (\%)} & \textbf{Mean queries} & \textbf{Median queries} \\ \hline
            (1+1)-ES & 6.22 & 9640.20 (4220.80) & $10^4$ \\
            NES & 4.24 & 9701.60 (2960.40) & $10^4$ \\
            CMA-ES & 77.09 & 5484.50  (4142.30) & 4876.00 \\ \hline
        \end{tabular}
    \caption{Comparison of the three evolutionary algorithms on the generation of targeted adversarial attacks for ResNet-50 pretrained on ImageNet. The values in parentheses represent the mean number of queries computed only on the successful attacks.}
    \label{tab:targeted-resnet-50}
\end{table}

Also in this case, we can observe that the performance of CMA-ES is significantly better than (1+1)-ES and NES.
Moreover, we can observe that in this setting the performance of (1+1)-ES seems to be comparable to that of NES.

%----------------------------------------------

\section{Conclusions}
\label{sec:conclusions}
Despite being very powerful in computer vision tasks, deep neural networks may be brittle to adversarial attacks.
Adversarial attacks may be easily carried out when the attacker has access to the gradients of the attacked neural network.
However, in most real-world attack scenarios the attacker does not have access to the gradients, so it must rely on black-box methodologies, such as those offered by evolutionary computation.

In this paper, we compared three different evolution strategies for the generation of black-box untargeted and targeted adversarial attacks: (1+1)-ES, Natural Evolution Strategies and CMA-ES.
We tested these algorithms on three well-known neural networks (VGG16, Inception-v3 and ResNet50) on a widely known computer vision benchmark, the ImageNet dataset.
The results show that all the algorithms were able to find untargeted adversarial attacks for the high majority of the samples in ImageNet.
Moreover, a more focused analysis on ResNet-50 revealed that that only CMA-ES was able to find good adversarial attacks with very small perturbations (intended as $L^\infty$-norm) of the input.
Another advantage of CMA-ES is that it required a lower number of mean queries to successfully attack the neural network in all the test cases.
These results suggest that CMA-ES is better than the other two algorithms at both exploring and exploiting the landscape of adversarial perturbation under the observed setup.
Finally, the experiments on the targeted attacks for the ResNet50 confirmed the superiority of CMA-ES as a generator of adversarial samples, since it showed a success rate more than an order of magnitude higher than the other two algorithms.
In fact, CMA-ES was able to successfully attack more than 3/4 of the images also in this setting, where the other two algorithms were extremely ineffective.

This study did not take into account the generation of adversarial samples for neural networks that included adversarial examples in the training set.
Therefore, in future work we plan to benchmark the three evolutionary algorithms on neural networks trained to either recognize or be robust to adversarial examples. Furthermore, we will extend the comparison to other optimization algorithms, in particular more recent variants of CMA-ES such as those presented in \cite{vermetten2019online, varelas2018comparative}. 
Finally, we will consider the generation of adversarial examples with constraints on other norms (single-objective optimization) or a combination of different norms (multi-objective optimization). %Rew 3.2

\bibliographystyle{ACM-Reference-Format}
\bibliography{main_gecco} 

\end{document}